\pdfoutput=1

\documentclass[11pt]{article}

\usepackage{authblk}

\usepackage[]{acl}

\usepackage{times}
\usepackage{latexsym}

\usepackage[T1]{fontenc}

\usepackage[utf8]{inputenc}

\usepackage{microtype}

\usepackage{inconsolata}

\usepackage{graphicx}

\usepackage{booktabs}

\title{Towards AI-assisted Academic Writing}

\author[1]{Daniel J. Liebling\textsuperscript{\textdagger}}
\author[1]{Malcolm Kane\textsuperscript{*}}
\author[2]{Madeleine Grunde-Mclaughlin}
\author[1]{\authorcr Ian J. Lang\textsuperscript{\ddag}}
\author[1]{Subhashini Venugopalan\textsuperscript{*}}
\author[1]{Michael P. Brenner\textsuperscript{\ddag}}
\affil[1]{Google Research, \textsuperscript{\textdagger}Seattle | \textsuperscript{*}Mountain View | \textsuperscript{\ddag}Cambridge, USA}
\affil[2]{Paul G. Allen School of Computer Science \& Engineering, University of Washington, Seattle, WA, USA}

\begin{document}
\maketitle
\begin{abstract}

We present components of an AI-assisted academic writing system including citation recommendation and introduction writing. The system recommends citations by considering the user’s current document context to provide relevant suggestions. It generates introductions in a structured fashion, situating the contributions of the research relative to prior work. We demonstrate the effectiveness of the components through quantitative evaluations. Finally, the paper presents qualitative research exploring how researchers incorporate citations into their writing workflows. Our findings indicate that there is demand for precise AI-assisted writing systems and simple, effective methods for meeting those needs.

\end{abstract}

\section{Introduction}

Scientific communication, including writing, is a necessary professional skill set. For example, The American Chemical Society guidelines for undergraduate education indicate that students must ``learn how to communicate technical information \ldots clearly and concisely, [i]n a scientifically appropriate style''' \cite{acsguidelines}. Writing effective prose is a skill developed through practice and feedback. Although the vast majority of scholarly publications are written in English, most English-language authors are not L1 English speakers. The proficiency gap negatively affects productivity of non-L1 authors. For example, \citet{flowerdew} found that over two-thirds of L1 Cantonese academic authors, writing in English, felt disadvantaged relative to L1 English speakers. Even for L1 speakers, the precise nature of scholarly language takes time and practice to develop expertise. \citet{morris2023scientists} interviewed scientists, who noted that their students were often ``not strong writers.'' The respondents anticipated that assisted writing would ``improve writing quality for a large number of students.'' 

Writing is fundamentally a task of translating ideas into text \cite{Flower1981ACP}. Interactive writing systems guide authors through the writing process by destructuring the writing process and, most recently, generating fluent language. Borrowing from crowdsourcing, \textit{Play Write} \cite{playwrite} ``selfsourced'' writing tasks through microtasks divorced from the document editor and delivered to the end user by an app. The tasks included outputs of typical NLP tasks such as summarization and grammar correction. Whereas older systems delegated tasks to the individual or crowd works, recent works incorporate LLMs as user-guided co-creators. For example, \textit{Wordcraft} \cite{wordcraft} focused on story writing. The system provided affordances for rewriting, elaborating, and open text generation. 
\textit{Sparks} \cite{sparks-in2writing,sparks-dis} used a LLM to suggest starter sentences intended to catalyze creative, compact writing for a general audience. Similar to users of Wordcraft, users of Sparks found value simply in generating narrative.

Computer-assisted writing is not a new concept; among others, \citet{mahlow2023writingtoolslookinglook} notes that AI-assisted writing is already commonplace. Modern LLMs are capable of generating text in scientific contexts comparable to expert human authors (e.g. \citet{wang-etal-2019-paperrobot,Ali2023PerformanceOC,gao_comparing_2023}) although this depends on the context (c.f. \citet{ruggeri-etal-2023-dataset}). Scientifically-grounded text generation is part of a larger adoption of AI in the sciences \cite{Hope2022ACI}. In this paper, we present two affordances for generative text in scientific contexts: citation recommendation and introduction writing. We develop and evaluate these affordances in the context of user-facing AI-assisted writing. Finally, we present the results of qualitative research on how researchers incorporate citation recommendations into their workflows. The system and findings show that AI-assisted writing is capable of generating useful content for academic authors, and that richer \textit{in situ} affordances can provide users with agency to craft more precise scholarly manuscripts.

\section{Contextual citation recommendation}
The citation recommendation task is typically framed as a recommender system that produces a ranked list of possible citations. Various approaches developed over time as machine learning methods evolved. \citet{farber_citation_2020} provide an overview of techniques that predate large language models (LLMs). Most techniques encode academic works into some semantic similarity (e.g. a topic model \cite{Kataria_Mitra_Bhatia_2010} or an embedding \cite{scibert}). Graph-based approaches \cite{10.1145/3640457.3688171} use the directed graph of citations as features or as a network for propagation of existing features.

Locating, copying, and formatting citations to include in the project takes time and effort. When performed concomitantly with writing, this context-switch between citation discovery can interrupt the user's writing flow.

We imagine \textit{in situ} citation recommendation as a task which recommends citations given the user's context and focus. Here, the focus is the cursor (insertion point) in the active document, representing the desired location of the suggestion. The context is some substring of the document leading up to the insertion point. We envisioned multiple scenarios for suggestions, depending on how much context the author has, and what type of output they desire. Consider the known-item refinding task where the author knows of a specific work and wishes to cite it. Frequently, the author can recall details about the work that they wish to cite \cite{known-item,re-finding}, although the known details might be incorrect. The author might recall these with a lower degree of precision (i.e. ``about 5--10 years ago'' or ``at an NLP conference'' or ``from Yamada Hanako's lab''). Finally, the author might not recall any of the indexing details of the paper. Instead they might remember a summary of the contribution. These incomplete or incorrect semantic cues to the underlying item are opportunities for the system to use additional context and world knowledge for recommending citations.

\subsection{Implementation}
\label{sec:citation-reco-impl}

Our system recommends citations from two sources. First, a user's writing project typically contains one or more files with citations expressed as structured content, e.g. BibTeX. Second, the system contains a local database of scholarly works: a copy of the OpenAlex corpus \cite{openalex}. Each record in this corpus includes the work's author(s), title, abstract, date, publication venue, citation count, and so forth. We used a language detection classifier to exclude works that appeared to be written in a language other than English. Because the mode of citation count is zero, we also excluded uncited works. Since the experiments documented herein were performed, the current implementation of the system retains recent uncited works in the database to allow them to be surfaced. After filtering, our database copy had 60.3 million rows out of the original 263.3 million rows.

As an interactive system, reducing response latency is critical to user perception and satisfaction. The system uses a highly scalable approximate nearest neighbor search \cite{scam} index for rapid retrieval of similar records in an embedding space. We chose the SPECTER2 embedding \cite{scirepeval}, a multi-format embedding developed specifically to represent scientific documents. SPECTER2 was trained on data from 23 different fields, not limited to computer science. SPECTER2 embeddings outperformed existing models on retrieval tasks. Our system concatenates each paper's title and abstract (if available in the OpenAlex record), projecting this text into the SPECTER2 embedding space. 

In addition to works available within the user's BibTeX files, the system needs to find novel candidates from the index that satisfy the user's intent. We implement this recommender as a Retrieval Augmented Generation (RAG) system \cite{Gao2023RetrievalAugmentedGF}. To retrieve a set of relevant citations, the system queries the index of existing works. Recall that the works are represented by a vector embedding of the title and abstract. The system takes advantage of LLMs observed behavior of ``hallucinating'' nonexistent facts or concepts \cite{10.1145/3571730}. Essentially, we prompt the LLM to fabricate a likely citation and then use that to find real citations. To do this, the system supplies the LLM with a prompt (see Appendix \ref{sec:app_fabrication_prompt}) containing the previous, current, and subsequent sentences from the user's content. The current sentence contains a special token which indicates to the system where in the sentence the citation is desired. The prompt instructs the LLM to fabricate the title and abstract of a paper that satisfies the user's context. Note that the system does not care if the LLM's generated citation exists. Rather, the fabricated citations are used as queries into the index of existing works. The fabricated title and abstract are embedded using the SPECTER2 model, which creates a vector used to query the nearest-neighbor index. As implemented, at most 10 nearest neighbors are returned. 

Although each result could be ranked by its distance to the query vector in embedding space, we apply an additional layer of scoring. Each result retrieved from the index is formatted into a new prompt (Appendix~\ref{sec:app_scoring_prompt}. These results are formatted as JSON objects. Each result is also given a unique, short hexadecimal string as a ``key'' property. Keys are constructed rather than using ordinal numbers (1,~2,~\ldots) or letters (A,~B,~\ldots) to avoid label bias \cite{reif-schwartz-2024-beyond}. Some LLMs also exhibit order bias \cite{shi2024judgingjudgessystematicstudy}; we did not evaluate this in our study. The prompt instructs the LLM to output the key that matches ``best citation to support [the] claim.'' Rather than using the key as output, the system runs model inference and collects the model's \textit{scores} for each of the keys in the input. A model's output score for each key is, to an approximation, the log probability of outputting that key to complete the input (prompt). The results are then ranked by their respective scores. Prompt inference was only run once for each item; no additional sampling of LLM output was performed.

We also implemented pairwise comparison to score suggestions. \citet{qin-etal-2024-large} showed that LLMs can be used to rank by presenting pairwise choices and having the LLM choose one of the items. This method differs from the scoring method described above. The model is prompted to choose the item from a pair of items that best matches the prompt. By combining pairwise ranks, one can determine a total ranking. By focusing the model's attention on a smaller number of targets, adverse effects from irrelevant targets are avoided.\footnote{c.f. \citet{rag-noise-good}, where noise improves quality.} Constructing the total ordering requires many pairwise comparisons. Although some techniques for reducing the quantity of comparisons exist \cite{bradley-rank-1952,pairwise}, we discarded this method due to the substantial increase in inference time, favoring the scoring method above.

The online citation recommender system allows the user to request a set of citation suggestions by right-clicking in the text editor. The client sends a substring of text adjacent to the insertion point, as well as the contents of BibTeX files. The latter include structured data about publications the author intends to cite.

\subsection{Evaluation}

To assess the efficacy of our citation recommendation system, we evaluated the LLM’s performance on the task of retrieving ground truth citations extracted from existing papers. The evaluation dataset was created from papers in S2ORC, a corpus of over 81 million papers spanning STEM disciplines \cite{lo-etal-2020-s2orc}. We uniformly sampled 0.1\% of papers from this corpus, then filtered to papers that include at least 10 sentences that include citations that existed in OpenAlex prior to September 2023 (our cutoff date). This ensured that the system would have access to titles and abstracts for these citations and would be able to use them as distractors in our evaluations. Five citation-containing sentences were randomly sampled from each qualifying paper, resulting in a dataset of 1015 sentences.

For each sentence, we gathered the necessary inputs to run the suggestion citation prompt described in Section~\ref{sec:citation-reco-impl}. This includes the target sentence's surrounding context and titles and abstracts of $n$ possible citations, for $n \in \{ 3, 5, 10 \}$. 

The $n$ candidate citations included the ground truth citation and $n-1$ distractor citations. We chose distractors in three different ways to test the system under varying difficulty. From least to most difficult, distractors were chosen uniformly randomly from: 

\begin{itemize}
    \item all papers in the evaluation dataset (sample of S2ORC)
    \item the ground-truth citation's nearest neighbors in SPECTER2 embedding space
    \item the references of the source paper containing the test sentence, excluding the ground truth reference
\end{itemize}

\begin{table*}
\centering
\begin{tabular}{c|c|c|ccc}
Distractor Type & $n$ & MRR & $p@1$ & $p@3$ & $p@5$ \\
\midrule
Random              & 3  & 0.755 & 0.612 &       &       \\
Random              & 5  & 0.549 & 0.333 & 0.665 &       \\
Random              & 10 & 0.320 & 0.124 & 0.348 & 0.500 \\
\midrule
Nearest neighbors   & 3  & 0.661 & 0.428 &       &       \\
Nearest neighbors   & 5  & 0.506 & 0.254 & 0.661 &       \\
Nearest neighbors   & 10 & 0.300 & 0.110 & 0.327 & 0.523 \\
\midrule
References          & 3  & 0.676 & 0.462 &       &       \\
References          & 5  & 0.496 & 0.261 & 0.641 &       \\
References          & 10 & 0.308 & 0.109 & 0.326 & 0.519 \\
\bottomrule
\end{tabular}
\caption{Retrieval metrics for 1,015 contextual citation retrieval cases with $n$ targets.}
\label{tab:citation_results}
\end{table*}

We employ Precision at \textit{k} (P@k) and mean reciprocal rank (MRR) as evaluation metrics. Because the randomly chosen set of distractors is domain agnostic, we expect a paper chosen from S2ORC at random to be unrelated to the test sentence. The two more difficult distractor sources include papers that are semantically related. In the \textit{nearest neighbors} condition, one of the distractors could be a reasonable substitute for the ground truth citation, particularly for well-known results. 

Table~\ref{tab:citation_results} shows the results. As expected, the ground truth citation tends to rank higher against randomly selected distractors when compared to distractors drawn from the semantic space or from the manuscript's references. However, the distractor source has less effect on precision. In a live system that uses this method, the user would need to choose from multiple suggestions rather than having the system propose only the top-ranked item.

\section{Writing introductions}
\subsection{Generating introductions}

We frame the introduction writing task as a mapping from the manuscript and references to a small number of paragraphs. The related work in the introduction should act like a microscope: canonical works coarsely orient the reader to a subfield; important recent works provide fine adjustment to the specific research track. Upon this foundation, the introduction builds the case for the specific contribution of the manuscript that follows. Our prompt chain follows this paradigm in three steps.

First, the system uses an LLM to identify novel claims from the author's manuscript relative to other works that the author cited. It assumes that the author already documented references in their BibTeX files at this time. For each reference, the system looks up the corresponding record in the OpenAlex database, retaining only those where a title and abstract are available. These references are split into two groups: \textit{canonical} and \textit{recent}. The \textit{canonical} references were published more than Y years ago while \textit{recent} were published within the last Y years. As in other systems, our system uses the title and abstract as a rough substitute for the work itself \cite{li-ouyang-2024-related}. To perform the relative comparison, the system then extracts paragraphs from the author's current work. Each paragraph is then combined with each of the references to form tuples of (paragraph, title, abstract). The prompt (Appendix~\ref{sec:app_claims_prompt}) acts as a binary classifier that confounds relevance and novelty. The LLM assess if the each paragraph's content is related to the abstract of the author's paper \textit{and} it is novel relative to the abstract a cited paper. The idea is to use this filter to find the work's novel contributions for incorporation into the introduction.

Each paragraph then receives one or more votes from the binary classifier. The system filters out paragraphs with low support. 
The remaining paragraphs, assumed to discuss novel results, are then passed to a simple summarization prompt (Appendix~\ref{sec:app_summarize_prompt}. Although current LLMs have long context lengths, at the time of our experiments, the token limit was smaller, and hence the (possibly many) novel paragraphs needed to be reduced into a shorter text. 

Finally, the system combines the canonical works, recent works, and summary of novel contributions into the written introduction section using the prompt in Appendix~\ref{sec:app_compose_prompt}. Example output of running the prompt chain on this submission is provided in Appendix~\ref{sec:app_generated_intro}.

\begin{figure*}[h]
  \includegraphics[width=\textwidth]{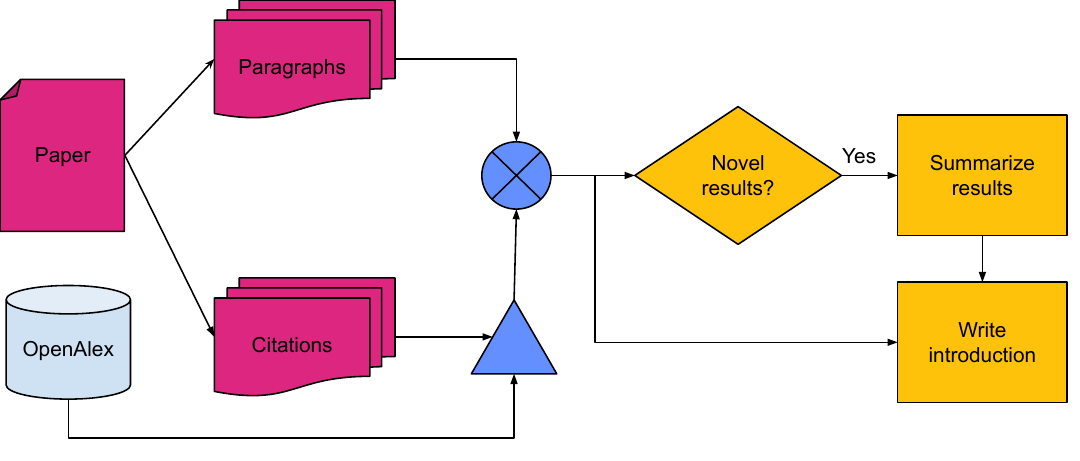}
  \caption{Flowchart from paper and citations to written introduction.}
\end{figure*}

\subsection{Evaluation}

\begin{figure}
    \centering
    \includegraphics[width=\columnwidth]{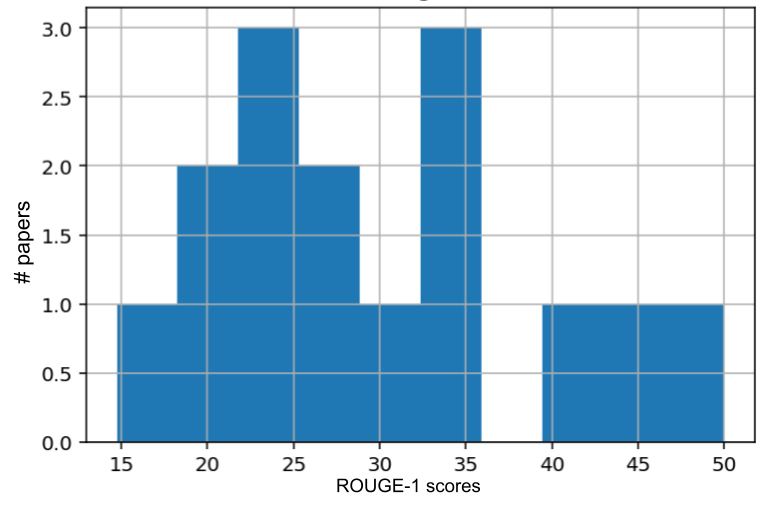}
    \caption{Distribution of ROUGE-1 scores for the generated introductions.}
    \label{fig:eval_nber_intro_rouge_dist}
\end{figure}

We evaluate the generated introductions using text metrics and by prompting an LLM. Our evaluation dataset is a subset of papers from the [United States] National Bureau of Economic Research\footnote{\url{https://www.nber.org/research/data}} (NBER). We extracted the introduction from 14 NBERs papers. For text evaluations, we use ROUGE~\cite{lin-2004-rouge} which is a recall-based metric and often used in the context of summarization. The average ROUGE score across the papers is 29.9, the distribution of scores is shown in Figure~\ref{fig:eval_nber_intro_rouge_dist}.

\begin{figure}
    \centering
    \includegraphics[width=\columnwidth]{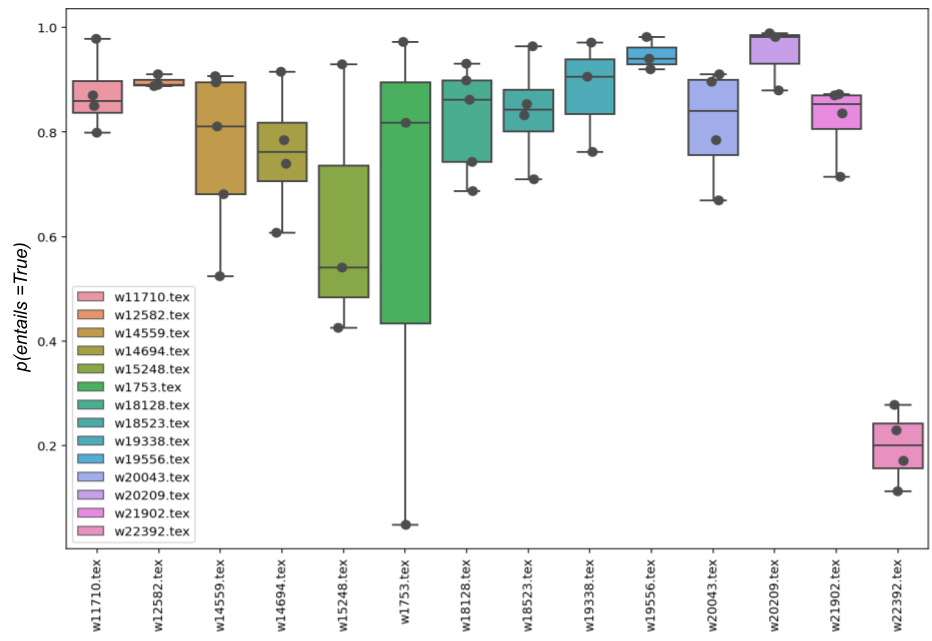}
    \caption{Distribution of scores for whether claims from the generated introductions \textit{entail} the original introduction based on an LLM.}
    \label{fig:eval_nber_intro_nli_score_dist}
\end{figure}

For LLM-based evaluations, we consider evaluating the introductions based on the claims made in the generated introductions in comparison to the original introductions. Specifically, we used a prompt (Appendix~\ref{sec:claim_extract_prompt}) to extract 3--5 claims from each of the generated texts. From 14 generated introductions, we extracted 52 claims. Then, we prompt the LLM to verify whether the claim from the generated introduction entails from the full original introduction. Appendix~\ref{sec:app_entailment_prompt} includes the full prompt used. We consider two versions of the LLM based evaluation: (1) we ask the LLM for a simple ``yes'' or ``no'' response for the prompt, (2) we consider the log-likelihood scores for the ``yes'' and ``no'' response tokens and normalize them to determine the probability that the generated claim is entailed from the the original. Figure~\ref{fig:eval_nber_intro_nli_score_dist} presents the probability scores for whether claims from the generated introduction entail from the original introduction. Of the 52 claims extracted from the generated introductions, 47 of them are entailed from the original introduction indicating a high degree of precision. In general, we find that the generated introductions score highly when the original introduction section hews closely to a single topic. Table~\ref{tab:intro_scores} compares a generated paragraph from NBER 20209 \cite{NBERw20209} with entailment score 0.983 versus a lower-performing paragraph from NBER 22392  \cite{NBERw22392} with entailment score 0.279. The higher-scoring generated passage captures the main concepts from the original work, while the lower-scoring passage hones in on natural disasters, which is not the theme of the original work. The LLM correctly identified the paragraph as not entailing the original. It seems reasonable to expect that self-critique \cite{NEURIPS2023_91edff07} can be used to improve generated introductions in the future.

\begin{table*}
\centering
\begin{tabular}{p{0.45\linewidth}|p{0.45\linewidth}}
Original excerpt & Generated excerpt \\
\midrule
\textit{NBER 20209, entailment 0.983} & \\
\small
It has been known, at least since the path-breaking work of \textcolor{red}{Arrow, that asset prices reflect a combination of stochastic discounting and probability distributions}. \ldots We demonstrate in Section 5 that a \textcolor{blue}{Perron–Frobenius approach} leads naturally to the construction of a martingale component \ldots & \small This research explores alternative solutions by investigating the application of \textcolor{blue}{Perron–Frobenius theory} to construct a \textcolor{red}{probability distribution from Arrow prices}. \ldots \\
\midrule
\textit{NBER 22392, entailment 0.279} & \\
\small
However, just because the inhabitants of some locations have higher incomes than others
does not mean there is a large causal effect of moving to these locations. \ldots Distinguishing between selection and direct causal effects of locations is challenging. Large, exogenous relocation shocks are few and far between \ldots We shed new light on the role of location in shaping economic outcomes by studying the consequences of a true “natural” experiment. On January 23, 1973, a \textcolor{purple}{long-dormant volcano erupted}
unexpectedly on the Westman Islands \ldots & \small The impact of natural disasters on economic development is a topic of considerable interest to economists. While the immediate consequences of \textcolor{purple}{natural disasters} are often well documented, the long-term impacts are less well understood. In particular, the impact on intergenerational mobility and the transmission of risk across generations is a critical yet underexplored area of research. <cit.> have shown that children who spend more time in a low-poverty environment have better long-term outcomes. This suggests that the effects of \textcolor{purple}{natural disasters} may not only be felt by those who experience them directly, but also by future generations.  \\
\bottomrule
\end{tabular}
\caption{Comparison of two generated introduction paragraphs with high and low entailment scores relative to the original text.}
\label{tab:intro_scores}
\end{table*}

\section{Opportunities}
As part of a study on citation verification, we conducted semi-structured interviews with academic authors about their experiences citing related work. Six researchers (5 self-identifying as men; 1 woman) from the research division of a large technology company participated. All researchers are experienced academic authors (mean \textit{h}-index 26, $\sigma=13$). Their research domains include sub-fields of computer science including quantum computing, virtual reality, biomedical imaging, natural language processing, and responsible artificial intelligence. Two of the six spoke English as a second language and all participants spoke at least one language other than English.

The semi-structured interviews covered the following topics: participants' current approaches to find and validate references, if their approaches would change with unlimited time and resources, how their approach differs depending on citation type, and imagined capabilities of an ideal support tool for citation verification and recommendation. We performed inductive thematic analysis of the interviewees’ statements. We performed three rounds of coding to create themes, resolving disagreements through conversation among two authors.

\textbf{Time constraints limit the validation process.} Nearly all participants raised the concern of careful validation. That is, they needed to understand specifically how the citation was relevant. However, several participants mentioned time constraints influencing their decision to cite works. Although every participant indicated that they sometimes cited papers that they had fully read, they also noted instances where they cited papers they had not entirely read. They employed skimming strategies while engaged in the literature review process in order to find more precisely related works.

Participants suggested various affordances for a tool to support the validation process. For example, one participant suggested finding the specific claims in the suggested citation that were related to the author’s citing text. Going to the original source was important because some participants remarked that papers’ claims can be misrepresented by citing authors, or the abstract did not accurately reflect the paper’s results. In interfaces for scholarly readers, existing systems such as \textit{Relatedly} \cite{relatedly} provide affordances similar to those suggested by the participants. The system we presented in this work only surfaces paper metadata such as title and abstract, so incorporating additional sensemaking affordances as part of the user’s workflow will support more rigorous citation suggestion.

\textbf{Surrounding text must be accurately scoped.} Participants also stressed the importance of having nuanced enough statements to accurately represent the paper [P1,P2]. They recognized that inaccurate corresponding text is often the consequences of human error or time constraints, rather than bad faith actions. Therefore, P2 expressed interest in support for rewriting existing text spans to better represent the cited paper. Our work finds a reference from a text span. Future work could also improve an existing span to better represent the reference. 

\textbf{Community norms impact reference choices.} Some participants felt pressure to cite ``the right'' source because peer reviewers would easily identify gaps in the related work. However, the precision of those citations varied depending on the field and relevance to the author’s work. Several suggested that the situating citations might be more interchangeable than the more recent works. 

\textbf{Importance of contextualization within the broader literature.} Participants reflected that although a given citation may be relevant, it may not be sufficient [P3,P4]. For instance, multiple references may be needed if the statement is multifaceted and nuanced, or if the statement is broad and requires a set of references. This idea of sufficiency extended to the reference set of entire sections, as P1 expressed concerns about misrepresenting sub-fields when merging or combining subsections of a related work. 

Part of the challenge of building a good reference set is understanding the broader trends of the overall field. Participants expressed interest in a tool that bridges relevant but separate streams of the literature, whether it be similar methods and theories from a different field or differing methods and theories from a similar field [P1,P2,P3]. The challenge of becoming aware of and fully encapsulating these different strands motivated their wish for a tool with a broad sense of the literature. These reflections suggest that reference selection must be valid on multiple levels, with each individual reference accurately represented in the close text and the set of references sufficient in representing the overall literature. Our tool focuses on the former, and there is a rich opportunity for future work in the latter.

\section{Conclusion}
As a highly developed, precise form of communication, the skill set of academic writing takes time to develop. The writing process requires focus, yet can be disrupted by related tasks such as the curation of related work. The qualitative research showed that even experienced authors have nuanced procedures for identifying and citing prior work. Rather than fully replace academic authors, it seems more likely that writing assistants will continue to proliferate, capturing a rich design space \cite{design-space}. In this paper, we presented two affordances for academic writing framed in the context of a live authoring experience: suggesting citations in the context of the document, and writing an introduction section. Quantitative evaluation shows that these methods are capable of generating content that augments the author's writing process.

\section{Limitations}

The system, studies, and participants described herein were only evaluated on English-language documents and queries, although five of the six participants were fluent in a language other than English. The OpenAlex corpus includes non-English documents, but we excluded those from our database. Finally, citation suggestion is an inherently biased task. Simple filters such as citation count prevent the discovery of ``sleeping beauties'' \cite{sleeping-beauty}, while heuristics such as the venue's impact factor may obscure novel ideas that have not made it into mainstream publication. Systems that take diverse viewpoints into account, and present them to authors in an interpretable fashion, will help diffuse novel ideas into scientific discourse.

\section{Acknowledgments}

We thank Paul Raccuglia for work that inspired this project, Eun-Ah Kim for feedback on situating introduction sections of scholarly manuscripts, and John Platt for reviewing early generative output. We acknowledge the contribution of open-source contributors to Python packages including but not limited to \texttt{pylatexenc}, \texttt{TexSoup}, \texttt{bibtexparser}, and \texttt{more\_itertools}.

\bibliography{custom}

\appendix

\clearpage
\onecolumn
\label{sec:appendix}
Prompts templates are processed using the Jinja\footnote{https://jinja.palletsprojects.com/en/stable/} templating library. Line breaks shown here may not match the line breaks used in the text prompt.

\section{Prompts for suggesting citations}
\subsection{Citation fabrication}
\label{sec:app_fabrication_prompt}
\begin{verbatim}
You are an expert at suggesting relevant scientific papers.

I will provide some sentences from a paper that I am writing. In the sentences,
I will place a token CITE-HERE where I need to cite a relevant paper. Your task
is to make up the title and abstract of a paper that you think would be relevant
to this context. Give your output in JSON format with values for keys "title"
and "abstract".

SENTENCES: {{ previous_sentence }} {{ masked_sentence }} {{ next_sentence }}

Now, make up the title and abstract of a paper that I should cite at the CITE-HERE token.

Answer:
\end{verbatim}

\subsection{Citation scoring}
\label{sec:app_scoring_prompt}
\begin{verbatim}
You are the editor at a prestigious scientific journal. The author of a paper asks
you to recommend the best citation to support their claim. You are given a set of
citations of papers in JSON format. Each citation includes a key in the "key" field,
the paper title in the "title" field, and the paper abstract in the "abstract" field.
You are also given an extraction of the paper, which indicates the location of the
desired citation with the string "CITE-HERE".

Select the best citation from the list of citations that best supports the context
of the extraction and give the value of the corresponding "key" field. Only give me
the value, nothing else.

EXTRACTION
{{ previous_sentence }} {{ masked_sentence }}  {{ next_sentence }}

CITATIONS
[{%
{
  'key': {{ c['key'] }},
  'title': {{ c['title'] }},
  'abstract': {{ c['abstract'] }},
}
{%

The key of the citation that best fits this extraction is:
\end{verbatim}

\clearpage
\section{Prompts for writing introductions}
\subsection{Determining claims}
\label{sec:app_claims_prompt}
\textit{Extracts claims from the author's manuscript and compares them with existing work.}

\begin{verbatim}
Your task is to determine if a paragraph from a scientific paper discusses
a novel result. You are given the abstract of the paper, abstract of related
paper, and a paragraph from the body of the paper. You answer YES if and only
if the paragraph's content is related to the abstract of this paper, and it
is novel relative to the abstracts of related papers.

ABSTRACT OF THIS PAPER
{{ abstract }}

ABSTRACT OF A RELATED PAPER
{{ ref_chunk[1].abstract }}

PARAGRAPH FROM THIS PAPER
{{ ref_chunk[0] }}

QUESTION
Q: Does the paragraph from this paper show a novel result worth mentioning
in the introduction? Respond YES or NO and explain your answer in one
sentence.
A:
\end{verbatim}

\subsection{Summarizing claims}
\label{sec:app_summarize_prompt}
\textit{Summarizes claims extracted using the previous prompt.}

Inputs: \texttt{novel\_results}, a list of text chunks from a paper.

\begin{verbatim}
You are a scientist writing up the results of your work. The following
paragraphs contain information about your results. Summarize the key
results in a few sentences.

{%
  {{ result | trim }}
{%

Now summarize the results in a few sentences.
\end{verbatim}

\subsection{Composing introduction}
\label{sec:app_compose_prompt}
\textit{Final step in the prompt chain to compose the introduction section.}
Inputs: 

\begin{tabular}{|l|l|}
\hline
\textbf{Field name} & \textbf{Description} \\ \hline
\texttt{title} & Manuscript title \\ 
\texttt{results} & Summary of experimental results \\ 
\texttt{[genesis\_references]} & List of canonical references \\
\texttt{[recent\_references]} & List of recent references \\
\hline
\end{tabular}

\begin{verbatim}
Given a list of related work, and the results of a paper, write the
introduction section for that paper. Refer to any of the REFERENCE
papers using the id in that REFERENCE.

PAPER TITLE: {{ title }}

FUNDAMENTAL PAPERS IN THIS FIELD:

{%
REFERENCE #{{ loop.index }}:
{%
{%
{%

RECENT RESULTS THAT THIS PAPER BUILDS ON:

{%
REFERENCE #{{ loop.index + len(genesis_references) }}:
{%
{%
{%

RESULTS: {{ results }}

Now write the paper introduction. Cite references from both the FUNDAMENTAL PAPERS and
the RECENT RESULTS. When you cite a reference, use the reference number in brackets.
Begin and end your introduction with three single quotes (''').
{%

INTRODUCTION:
\end{verbatim}

\subsection{Claim extraction (for evaluation)}
\label{sec:claim_extract_prompt}
Inputs: \texttt{num\_claims}, the number of claims to extract from \texttt{introduction}, the text of the introduction.

\begin{verbatim}
Your job is to answer questions about the following introduction section of an
academic paper:

{{introduction}}

List {{num_claims}} novel claims that the introduction section makes. Output only
one sentence per claim and do not add additional explanation. Do not add any additional
formatting. Begin each claim with a number.
\end{verbatim}

\subsection{Entailment (for evaluation)}
This prompt was used for evaluation.
\label{sec:app_entailment_prompt}
\begin{verbatim}
You will be given a generated paragraph (hypothesis) and the text corresponding
to the original introduction (context). Your goal is to evaluate whether the
generated paragraph (hypothesis) entails the original introduction.

It is alright if the generated paragraph does not encompass the core of the
original introduction, as long as the generated paragraph does not introduce
any new claims.

Start your answer with "yes" or "no" depending on whether the generated
paragraph entails the original introduction and then provide a brief 1-2 
sentence explanation for why.

GENERATED PARAGRAPH (hypothesis):
{{gen_intro_para}}

Original introduction (context):
{{orig_intro}}

Use the following format:
<yes or no depending on whether hypothesis entails the context>

Explanation:
\end{verbatim}

\clearpage
\section{Generated output}
\label{sec:app_generated_intro}
\textit{This \textbf{AI-generated text} is the system's output using the introduction writing method, applied to this manuscript. The system did not have access to the human-authored introduction at the time of generation.}

The proliferation of academic literature presents challenges for researchers seeking to efficiently locate and incorporate relevant prior work into their writing.  Citation recommendation systems aim to alleviate this burden by suggesting relevant citations given a user's writing context.  Early approaches to this task relied on representing academic works using topic models \cite{Kataria_Mitra_Bhatia_2010} or embedding techniques \cite{scibert}, while graph-based methods leveraged citation networks to identify relevant work \cite{10.1145/3640457.3688171}. While these methods can identify broadly relevant papers, they often fail to capture the nuanced ways in which researchers incorporate citations within their writing.

A key challenge in citation recommendation is the need for systems to understand not only the topical relevance of a citation, but also its specific role within the author's argument. \citet{farber_citation_2020} provide a comprehensive overview of pre-LLM techniques for citation recommendation, highlighting the limitations of existing methods in capturing this contextual information. More recent work has explored the use of large language models (LLMs) for citation recommendation, leveraging their ability to process and understand natural language \cite{Gao2023RetrievalAugmentedGF}. These LLM-based approaches have shown promise in improving the accuracy and relevance of citation recommendations.

In this paper, we present a novel approach to AI-assisted academic writing that combines contextual citation recommendation with automated introduction generation. Our system leverages the power of LLMs to provide researchers with in-situ citation suggestions, recommending citations that are both topically relevant and appropriate for the specific context of their writing. 

We demonstrate the effectiveness of our approach through a quantitative evaluation on a dataset of academic papers. Our results show that our system is able to accurately identify relevant citations, even when presented with limited context.  Furthermore, we find that our system is able to generate coherent and informative introductions that accurately reflect the key contributions of a research paper. 

Beyond simply identifying relevant citations, our system is designed to support the broader process of academic writing. Recognizing that researchers often struggle to synthesize and integrate prior work into their own writing, we incorporate a novel introduction generation component that automatically generates structured introductions, situating the contributions of a research paper within the context of prior work. This component leverages LLMs to identify the novel contributions of a paper, summarize related work, and compose a coherent introduction that effectively sets the stage for the research presented in the manuscript.

\end{document}